\documentclass{article}
\usepackage{ijcai22}

\usepackage{times}

\usepackage{soul}
\usepackage{url}
\usepackage[hidelinks]{hyperref}
\usepackage[utf8]{inputenc}
\usepackage[small]{caption}
\usepackage{graphicx}
\usepackage{amsmath}
\usepackage{amssymb}
\usepackage{booktabs}
\usepackage{xcolor}
\usepackage{multirow}
\usepackage{placeins}
\usepackage{bm}
\usepackage{caption}
\usepackage{subcaption}

\DeclareMathOperator{\bbox}{R}
\DeclareMathOperator{\gtbbox}{R^G}
\DeclareMathOperator{\dtbbox}{R^D}
\DeclareMathOperator{\dtbboxalt}{\tilde{R}^D}
\DeclareMathOperator{\gtbboxalt}{\tilde{R}^G}
\DeclareMathOperator{\allgtbbox}{\mathcal{G}}
\DeclareMathOperator{\alldtbbox}{\mathcal{D}}
\DeclareMathOperator{\fdtbbox}{D(c)}
\DeclareMathOperator{\ped}{person}
\DeclareMathOperator{\iou}{IoU}
\DeclareMathOperator{\fn}{FN^G(c)}
\DeclareMathOperator{\fp}{FP^D(c)}
\DeclareMathOperator{\tp}{TP^G(c)}

\DeclareMathOperator{\fppi}{FPPI(c)}
\DeclareMathOperator{\gdpi}{GDPI}
\DeclareMathOperator{\lamr}{LAMR}
\DeclareMathOperator{\lamrR}{LAMR_{r}}
\DeclareMathOperator{\flamr}{FLAMR_{\errorpop}}
\DeclareMathOperator{\mrS}{MR}
\DeclareMathOperator{\fppiS}{FPPI}
\DeclareMathOperator{\sem}{\mathfrak{S}}
\DeclareMathOperator{\inst}{\mathfrak{I}}

\DeclareMathOperator{\errorpop}{P}
\DeclareMathOperator{\mrp}{MR_{\errorpop}(c)}

\DeclareMathOperator{\epopa}{\tilde{\mathcal{E}}}
\DeclareMathOperator{\epopb}{\mathcal{E}}

\DeclareMathOperator{\occlclasses}{O}
\DeclareMathOperator{\crpopa}{\tilde{\mathcal{C}}}
\DeclareMathOperator{\crpopb}{\mathcal{C}}

\DeclareMathOperator{\ambpop}{\mathcal{A}}

\DeclareMathOperator{\vispop}{\mathcal{V}}
\DeclareMathOperator{\occlpop}{\mathcal{O}}
\DeclareMathOperator{\potocclpop}{\tilde{\mathcal{O}}}
\DeclareMathOperator{\forepop}{\mathcal{F}}
\DeclareMathOperator{\fore}{F(c)}
\DeclareMathOperator{\backpop}{\mathcal{B}}
\DeclareMathOperator{\back}{B(c)}

\DeclareMathOperator{\cx}{cX}
\DeclareMathOperator{\cy}{cY}
\DeclareMathOperator{\caligned}{D}
\DeclareMathOperator{\spop}{\mathcal{S}}
\DeclareMathOperator{\s}{S(c)}
\DeclareMathOperator{\lpop}{\mathcal{L}}
\DeclareMathOperator{\lc}{L(c)}
\DeclareMathOperator{\gpop}{\mathcal{H}}
\DeclareMathOperator{\g}{H(c)}

\DeclareMathOperator{\igng}{\mathcal{I}^G}
\DeclareMathOperator{\ignd}{\mathcal{I}^D}

\DeclareMathOperator*{\argmax}{arg\,max}
\DeclareMathOperator*{\argmin}{arg\,min}
\DeclareMathOperator*{\chset}{C^{\gpop}}
\DeclareMathOperator*{\cset}{C}

\DeclareMathOperator{\flamrh}{FLAMR_{\errorpop}^{\gpop}}
\DeclareMathOperator{\flamrhf}{FLAMR_{\forepop}^{\gpop}}
\DeclareMathOperator{\flamrf}{FLAMR_{\forepop}}

\definecolor{red}{RGB}{255,0,0}
\definecolor{gray}{RGB}{128,128,128}
\definecolor{magenta}{RGB}{255,0,255}
\definecolor{orange}{RGB}{255,165,0}
\definecolor{cyan}{RGB}{0,255,255}
\definecolor{green}{RGB}{0,255,0}
\definecolor{blue}{RGB}{0,0,255}
\definecolor{brown}{RGB}{139,69,19}

\newcommand\blfootnote[1]{
    \begingroup
    \let\thefootnote\relax\footnotetext{\hspace{-16pt}#1}
    \endgroup
}

\title{Revisiting the Evaluation of Deep Neural Networks for Pedestrian Detection}

\author{
Patrick Feifel$^{1,2,}$\footnote{Equal contribution}\and
Benedikt Franke$^{3,4,*}$\and
Arne Raulf$^3$\and
Friedhelm Schwenker$^4$\and
Frank Bonarens$^{1}$\And
Frank Köster$^{3, 2}$
\affiliations
$^1$Stellantis, Opel Automobile GmbH\\
$^2$Carl von Ossietzky Universität Oldenburg\\
$^3$Deutsches Zentrum für Luft- und Raumfahrt\\
$^4$Universität Ulm\\
\emails
patrick.feifel@external.stellantis.com,
benedikt.franke@dlr.de
}

\begin{document}
\maketitle

\begin{abstract}
The reliable DNN-based perception of pedestrians represents a crucial step towards automated driving systems.
Currently applied metrics for a subset-based evaluation prohibit an application-oriented performance evaluation of DNNs for pedestrian detection.
We argue that the current limitation in evaluation can be mitigated by the use of image segmentation. 
In this work, we leverage the instance and semantic segmentation of Cityscapes to describe a rule-based categorization of potential detection errors for CityPersons. 
Based on our systematic categorization, the filtered log-average miss rate as a new performance metric for pedestrian detection is introduced.
Additionally, we derive and analyze a meaningful upper bound for the confidence threshold.
We train and evaluate four backbones as part of a generic pedestrian detector and achieve state-of-the-art performance on CityPersons by using a rather simple architecture.
Our results and comprehensible analysis show benefits of the newly proposed performance metrics. 
Code for evaluation is available at \href{https://github.com/BeFranke/ErrorCategoriesPedestrianDetection}
{https://github.com/BeFranke/ErrorCategories}.
\end{abstract}

\section{Introduction}
\label{sec:intro}

\blfootnote{Copyright © 2022 for this paper by its authors. Use permitted under Creative Commons License Attribution 4.0 International (CC BY 4.0).}

Pedestrian detection is a crucial perception task for automated driving systems (ADS). 
Due to high complexity of the ADS environment, supervised machine learning models such as deep neural networks (DNNs) outperform traditional computer vision models and meet the high performance standards. 
Hence, traditional methods such as HOG \cite{hog1} have been replaced by DNNs, which can be designed single-staged and anchor-free \cite{csp,zhang_attribute-aware_2020} or two-staged and anchor-based \cite{khan2022f2dnet}.\looseness=-1

\begin{figure}[t]
\centering
\includegraphics[width=\linewidth]{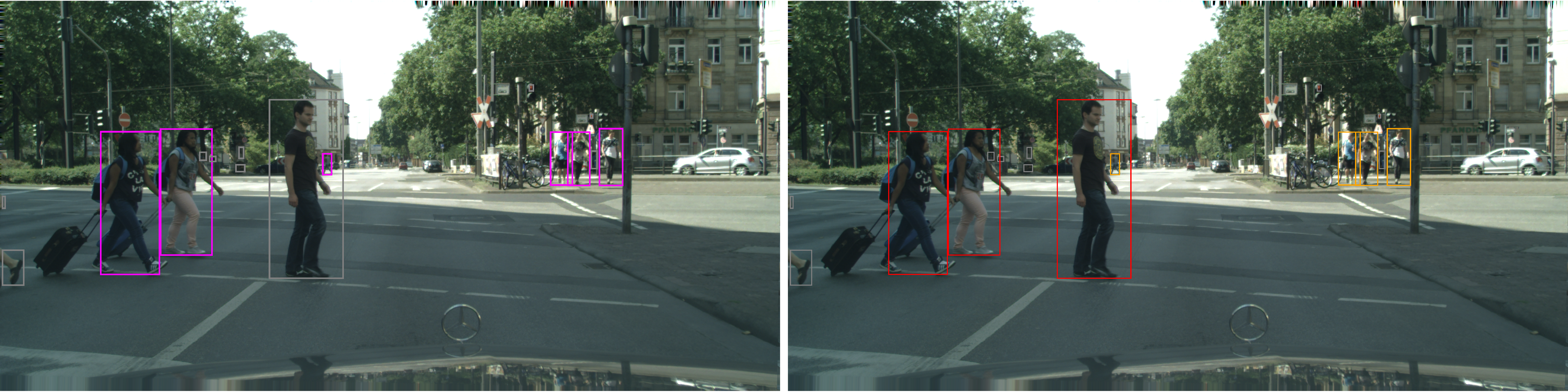}
\caption{
State-of-the-art DNNs for pedestrian detection are benchmarked with the log-average miss rate on the \textcolor{magenta}{reasonable} subset of the CityPersons validation dataset (left). 
From a safety perspective, particularly safety-critical pedestrians, such as the one standing directly in front of the automated vehicle, must be included in the evaluation and not be \textcolor{gray}{ignored}. 
Our proposed error categories (right) correctly distinguish between \textcolor{red}{foreground} and \textcolor{orange}{background}, among others. Based on them, we perform an application-oriented performance evaluation of DNNs for pedestrian detection. 
}
\label{fig:teaser}
\end{figure}

Avoiding false negatives is the main objective for pedestrian detection in an ADS.
A critical scene as shown in Figure \ref{fig:teaser} outlines the key task: A group of pedestrians cross the street right in front of the automated vehicle (AV). 
Intuitively, the evaluation should focus on pedestrians in the immediate vicinity of an AV, rather than distant pedestrians standing on the sidewalk in the middle of a crowd. 
The goal is to build relevant subsets of an evaluation dataset that contains these highly safety-critical pedestrians.
This enables a more meaningful performance evaluation of DNNs. 
Motivated by the Caltech evaluation protocol \cite{caltech}, occlusion-related or height-based subsets were proposed \cite{localization-error,repulsion-loss,occlusion,foreground}.\looseness=-1

The reasonable subset is most commonly used to benchmark DNNs for pedestrian detection. 
It is based on the visibility and pixel height of ground truth bounding boxes. 
As shown in Figure \ref{fig:teaser}, using this sparse information can result in particularly safety-critical pedestrians being ignored in the reasonable subset. 
As a consequence, currently used metrics only give limited information on the application-oriented performance.


Despite efforts to address highly occluded and therefore very difficult pedestrian detection cases, we argue that a realistic performance evaluation of a DNN for pedestrian detection should primarily address pedestrians in the near field of an AV.
In this sense, we think that a missed pedestrian in a distant crowd is less significant than a missed pedestrian standing directly in front of the AV.


Although a high recall is the primary objective for a DNN in a safety-critical application, the precision strongly influences the ADS operation in a complex environment. 
The current subset-based evaluation neglects the impact of different forms of false positives. 
We argue that multiple detections of the same pedestrian are less problematic for an ADS than false positives randomly scattered in the scene without reference to pedestrian-like features. 
Thus, a clear distinction between false positives must be found. 


In our work, we introduce a systematically derived categorization of errors that can leverage a safety argumentation for the DNN-based perception of pedestrians. 
Our contribution can be summarized as follows:

\begin{enumerate}
\item We propose a rule-based categorization that describes potential errors of a DNN for pedestrian detection.
\item We define novel performance metrics focusing on safety-critical pedestrians that enable application-oriented DNN evaluation.
\item We report results and analyze 44 different DNNs for pedestrian detection, divided into 11 training runs for four backbones.

\end{enumerate}

\section{Background}

Pedestrian detection is usually done by locating a 2d bounding box and assigning the correct class.
Most commonly, DNNs for pedestrian detection are evaluated with the log-average miss rate ($\lamr$) \cite{dollar_pedestrian_2011} on the \textit{reasonable} subset of the CityPersons \cite{citypersons} validation dataset. 
We refer to this performance metric as $\lamrR$. 
Since pedestrian detection is highly safety-critical and relevant to an ADS, the $\lamrR$ aggregates the miss rate (MR) and false positives per image ($\fppiS$). 

\begin{table}[ht]
\centering
\begin{tabular}{l|c|ccc}
\toprule
Method & $R$ & $B$ & $P$ & $H$ \\
\midrule
\midrule
CSP~\shortcite{csp}     & 11.0 & 7.3 & 10.4 & 49.3 \\
NOH-NMS~\shortcite{zhou2020noh}         & 10.8 & 6.6 & 11.2 & 53.0 \\
RepLoss~\shortcite{repulsion-loss}      & 10.9 & 6.3 & 13.4 & 52.9 \\
PRNet~\shortcite{prnet}                 & 10.8 & 6.8 & 10.0 & 53.3 \\
Beta R-CNN~\shortcite{betarcnn}         & 10.6 & 6.4 & 10.3 & 47.1 \\
NMS-Loss~\shortcite{nmsloss}            & 10.1 & - & - & - \\
Cascade R-CNN~\shortcite{cascadercnn}   & 9.2 & - & - & 36.9\\    
BGCNet~\shortcite{li_box_2020}          & 8.8 & 6.1 & 8.0 & 43.9 \\
APD~\shortcite{zhang_attribute-aware_2020} & 8.8 & 5.8 & 8.3 & 46.6 \\
F2DNet~\shortcite{khan2022f2dnet}       & \textbf{8.7} & - & - & 32.6 \\
\bottomrule
\end{tabular}
\caption{LAMR [\%] for different subsets of the CityPersons validation dataset: reasonable ($R$), bare ($B$), partial ($P$) and heavy ($H$).}
\label{tab:lamr_literature}
\end{table}


Table \ref{tab:lamr_literature} gives an overview of state-of-the-art DNNs for pedestrian detection that are evaluated on different subsets of CityPersons. 
The definitions of the subsets are based on the height interval $h=[50, 1024]$ and a varying visibility range $\upsilon = \frac{|\gtbbox_{vis}|}{|\gtbbox|}$ of a pedestrian: reasonable ($\upsilon=[0.65, 1]$), bare ($\upsilon=[0.90, 1]$), partial ($\upsilon=[0.65, 0.90]$) and heavy ($\upsilon=[0, 0.65]$).

\begin{figure*}[ht]
\centering
\includegraphics[width=\linewidth]{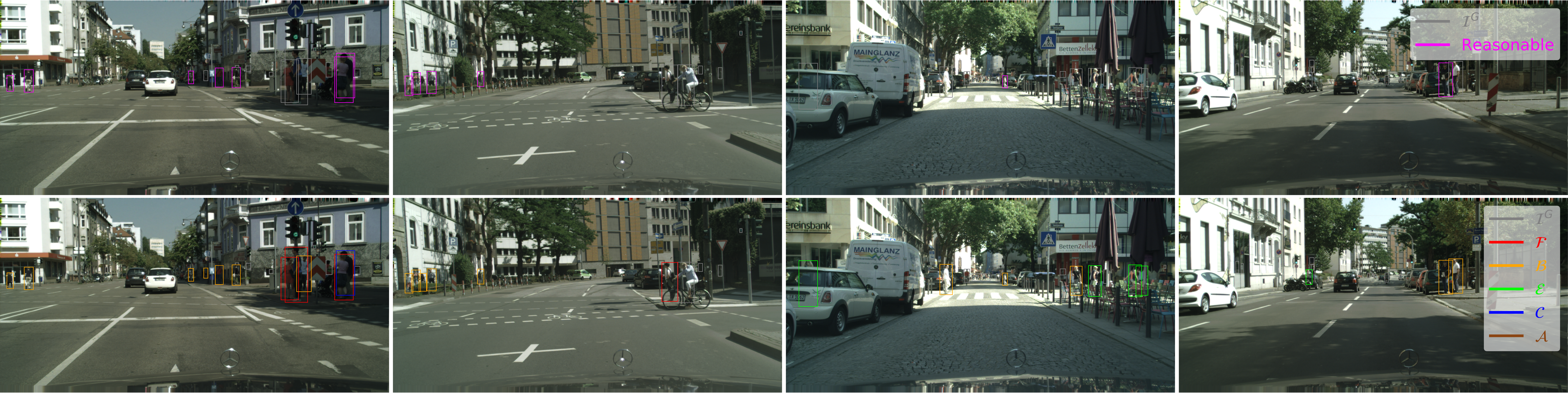}
\caption{
Incorrectly \textcolor{gray}{ignored} bounding boxes from the \textcolor{magenta}{reasonable} subset of CityPersons are recovered by our proposed error categories.}
\label{fig:re_vs_err_cat_gt}
\end{figure*}

\section{Generic Pedestrian Detector}

In this work, we provide a comprehensive analysis of different \textit{backbones} that are commonly used for DNNs for pedestrian detection. 
To achieve comparable results, we propose a DNN-based and generic pedestrian detector (GPD) consisting of \textit{feature extraction} and \textit{perception heads}. 

\paragraph{Feature Extraction}

Pre-trained image classification networks form the backbone of the feature extraction. 
To utilize backbones for pedestrian detection, additional layers (ALs) must be implemented. 
Based on computed features for various scales by the backbone, the feature extraction outputs a representation for a given input image.
In our work, we use the following feature extractions:

\begin{itemize}
\item \textbf{CSP-ResNet-50}: CSP \cite{csp} creates high-level semantic features based on ResNet-50 \cite{resnet} and deconvolutions.
\item \textbf{FPN-ResNet-50}: Feature pyramid network (FPN) \cite{zhang_attribute-aware_2020} that adds a pyramidal decoder to ResNet-50 to combine features from different scales.
\item \textbf{MDLA-UP-34}: Modified DLA (MDLA) \cite{zhou_objects_2019} augmentes DLA-34 \cite{dla} with deformable convolutions from lower layers to the output.
\item \textbf{BGC-HRNet-w32}: BGC \cite{li_box_2020} adds deconvolutions to a HRNet-w32 \cite{hrnet} concatenating the outputs to form the final representation.
\end{itemize}

\paragraph{Perception Heads}

In total, we have three perception heads taking extracted features as inputs and outputting a center, scale (height w/o width) and offset map.
Similar to APD \cite{zhang_attribute-aware_2020}, we apply 3x3 convolutions for each perception head.

\paragraph{Training}

We train and evaluate different GPD instances with varying pre-trained backbones on the CityPersons dataset.
In the following, an instance of GPD is simply referred to as a pedestrian detector (PD).
All PDs are trained with the Adam optimizer \cite{adam} without weight decay and a reduced image size of 640x1028 pixels. 
A linear warm up strategy is employed that increases the learning rate from $5 \cdot 10^{-8}$ to the final learning rate of $10^{-4}$ over 2000 iterations. 
We train for a maximum of 50k iterations on 2 GPUs with a batch size of 8.
The final PD is given by the best checkpoint with the lowest LAMR score on the reasonable subset of the CityPersons validation dataset. 
ResNet-50\footnote{\url{https://pytorch.org/hub/pytorch_vision_resnet/}}, DLA-34\footnote{\url{http://dl.yf.io/dla/models/imagenet/dla34-ba72cf86.pth}} and HRNet-w32\footnote{\url{https://github.com/HRNet/HRNet-Image-Classification}} are used as pre-trained backbones on ImageNet. 
Furthermore, we apply the center, scale and offset loss terms according to CSP \cite{csp}. Common data augmentation techniques like modifying brightness, contrast or saturation are applied.

\paragraph{Inference}
For post-processing, we apply a confidence threshold of 0.01 and use NMS with a threshold of 0.5.
The inference of PDs is conducted with the original image size of 1024x2048 pixels. 
Ground truth and detection bounding boxes are clipped to the image size.

\section{Methodology}

\begin{figure}[ht]
\centering
\includegraphics[width=\linewidth]{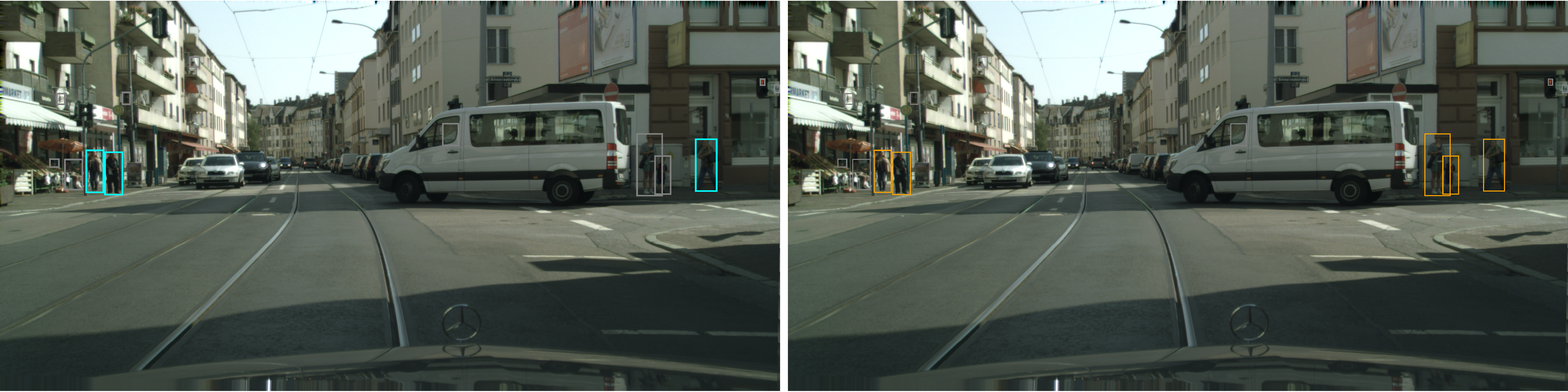}
\caption{Incorrectly \textcolor{gray}{ignored} bounding boxes from the \textcolor{cyan}{bare} subset of CityPersons are re-grouped to \textcolor{orange}{background}.}
\label{fig:ba_vs_err_cat_gt}
\end{figure}

In the following, we introduce different categories for ground truth bounding boxes $\gtbbox$ of the CityPersons validation dataset. 
Matching ground truth with detection bounding boxes $\dtbbox$ based on our systematic categorization identifies errors for false negatives and false positives.
Reducing false negatives is the primary safety-related objective during PD training. 
Intuitively, we expect false negatives to be positively correlated with pedestrian occlusion by other pedestrians and other environmental objects.
That's why categories regarding false negatives build upon the description of different forms of occlusions.
As shown in Figure \ref{fig:re_vs_err_cat_gt} and Figure \ref{fig:ba_vs_err_cat_gt}, our proposed error categorization recovers ignored pedestrians for the reasonable and bare subset of the CityPersons validation dataset.
Finally, we categorize false positives to identify the most disruptive ones for an ADS.

\label{sec:methodology}

\paragraph{Bounding Boxes}

We define a bounding box $\bbox$ as set of all pixels with $(x, y)$ corner coordinates that fall into the bounding box: $\bbox = \{ (x, y) \mid x_{1} \le x < x_{2} \land y_{1} \le y < y_{2} \}$.
Therefore, the width of the bounding box is defined as $w(\bbox) = x_2 - x_1$ and similarly the height $h(\bbox) = y_2 - y_1$.
A ground truth bounding box is denoted as $\gtbbox \in \allgtbbox$ in contrast to a detection bounding box $R^{\tilde{D}} \in \tilde{\mathcal{D}}$ which is associated with a confidence score $p(\dtbbox)$. 
Because of highly overlapping detections, post-processing methods such as non-maximum suppression $( \text{NMS}: \tilde{\mathcal{D}} \rightarrow \mathcal{D} )$ are applied to reduce the number of detections to $\dtbbox \in \mathcal{D}$. 
Based on a predefined confidence threshold $c$, less confident detections are ignored: 
$\fdtbbox = \{ \dtbbox \mid \dtbbox \in \alldtbbox \;\land\; p(\dtbbox) > c \}$.

Generally, a pixel-precise match between bounding boxes can not be expected.
Therefore the intersection over union (IoU) is used to measure the localization quality of $\dtbbox$ for $\gtbbox$.
The set of true positives is defined as:

\begin{equation}
\label{eq:tp}
\begin{split}
    \tp = & \{ \gtbbox \mid \gtbbox \in \allgtbbox \;\land\; \exists \dtbbox \in \fdtbbox: \\
    & \big [ \iou(\gtbbox, \dtbbox) > 0.5 \land \nexists \gtbboxalt \in \allgtbbox: \\
    & \iou(\gtbboxalt, \dtbbox) > \iou(\gtbbox, \dtbbox) \big ]\}
\end{split}
\end{equation}

A ground truth bounding box $\gtbbox$ that can not be matched is a false negative $\fn = \allgtbbox \setminus \tp$.
A detection bounding box $\dtbbox$ that can not be matched or can only be matched to an already matched $\gtbbox$ is assigned to the set of false positives:

\begin{equation}
\begin{split}
    \fp = & \{ \dtbbox \mid \dtbbox \in \fdtbbox \;\land\; \nexists \gtbbox \in \allgtbbox:  \\
    & \iou(\gtbbox, \dtbbox) > 0.5 \;\lor \exists \gtbbox \in \allgtbbox: \\
    & \big[ \iou(\gtbbox, \dtbbox) > 0.5 \;\land\; \exists \dtbboxalt \in \alldtbbox: \\
    & \iou(\gtbbox, \dtbboxalt) > \iou(\gtbbox, \dtbbox) \big] \}
\end{split}
\end{equation}

\paragraph{Image Segmentation}

In this work, we employ ground truth for semantic segmentation $\sem$ and instance segmentation $\inst$ to refine the subset-based evaluation of DNNs for pedestrian detection.
$\sem[x, y] = \ped$ means that the pixel at position $(x, y)$ belongs to a pedestrian. 
$\inst[x, y] = i$ means that the pixel at position $(x, y)$ has the instance ID $i$. 

\subsection{Error Categories for False Negatives}
\label{sec:error-categories}

\begin{figure*}[ht]
\centering
\includegraphics[width=\linewidth]{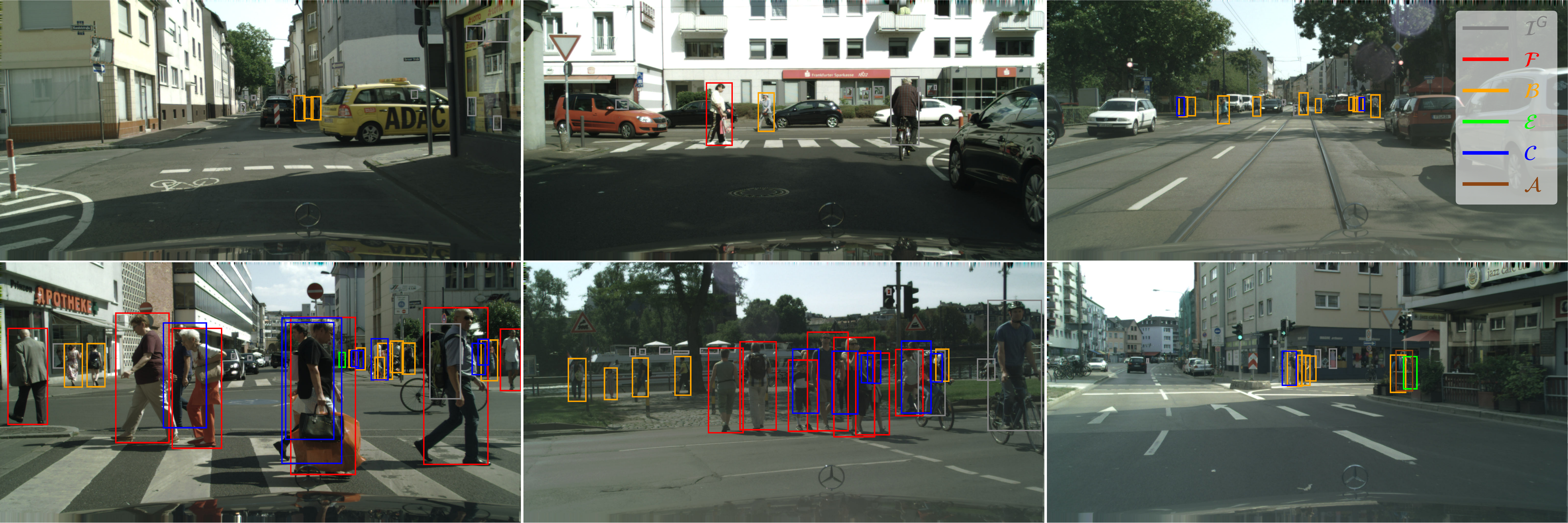}
\caption{Categories for ground truth bounding boxes:
\textcolor{red}{foreground $\forepop$},
\textcolor{orange}{background $\backpop$},
\textcolor{green}{environmental occlusion $\epopb$},
\textcolor{blue}{crowd occlusion $\crpopb$},
\textcolor{brown}{ambiguous occlusion $\ambpop$}.
\textcolor{gray}{Ignored} bounding boxes \textcolor{gray}{$\igng$} are not part of the evaluation.}
\label{fig:gt_err_cat_gt}
\end{figure*}

We define five error categories that separate ground truth bounding boxes of occluded pedestrians as well as highly safety-relevant pedestrians standing in the foreground or background.
Examples of our categorization are shown in Figure \ref{fig:gt_err_cat_gt}.
We propose a two-stage process to detect occlusion. 
First, potentially occluded pedestrians are identified based on the segmentation-based visibility $\phi$, where $i$ represents the instance ID belonging to the pedestrian: 

\begin{equation}
    \phi(\gtbbox, i) = \frac{| \gtbbox \;\cap\; \{ (x, y)\mid \inst[x, y] = i \} |}{|\gtbbox|}
\end{equation}

The set of occlusion candidates $\potocclpop$ builds upon the threshold $\lambda_{\phi}$: $\potocclpop = \{ \gtbbox \mid \gtbbox \in \allgtbbox \;\land\; \phi_{c} (\gtbbox) < \lambda_{\phi} \}$.
For our experiments, we empirically set  $\lambda_\phi = 0.6$.

\paragraph{Environmental Occlusion}

Environmental occlusion occurs when a pedestrian is partially hidden behind objects in the scene, e.g. traffic signs, vegetation or cars. 
We define $\occlclasses$ as 20 selected classes of the semantic segmentation $\sem$ of Cityscapes \cite{cityscapes} that can potentially cause occlusion. 
Truncated bounding boxes belong to this category, as the area that extends beyond the image is understood as environmental occlusion. 
We define the visibility with respect to the environment $\phi_{e}$ as

\begin{equation}
    \phi_{e} (\gtbbox) = \frac{| \gtbbox \;\cap\; \{ (x, y) \mid \sem[x, y] \in \occlclasses\}|}
    {|\gtbbox|}
\end{equation}

For our experiments, we empirically set $\lambda_e = 0.7$. We define the intermediate set of environmentally occluded ground truth bounding boxes as $\epopa = \{ \gtbbox \mid \gtbbox \in \potocclpop \;\land\; \phi_{e} (\gtbbox) > \lambda_{e} \}$.

\paragraph{Crowd Occlusion}

Crowd occlusion (also intra-class occlusion \cite{repulsion-loss}) occurs when a pedestrian is occluded by other pedestrians. We define the intra-class visibility $\phi_{c}$ that describes the relation of the instance area of a pedestrian to the semantic area occupied by the person class:

\begin{equation}
    \phi_{c} (\gtbbox, i) = \frac{| \gtbbox \;\cap\; \{ (x, y)\mid \inst[x, y] = i \} |}
    {| \gtbbox \;\cap\; \{ (x, y)\mid \sem[x, y] = \ped\}|}
\end{equation}

We introduce the threshold $\lambda_{c}$ and define the intermediate set of crowd occluded ground truth bounding boxes as $\crpopa = \{ \gtbbox \mid\gtbbox \in \potocclpop \;\land\;  \phi_{c} (\gtbbox, i) > \lambda_{c} \}$.
For our experiments, we empirically set $\lambda_c = 0.5$.

\paragraph{Ambiguous Occlusion}

Ambiguous occlusion occurs when pedestrians are simultaneously occluded by the environment and other pedestrians. We introduce the ambiguity factor $\lambda_{a} \in (0, 1)$ to relax the thresholds for crowd and environment occlusion and define
$\ambpop^E = \{ \gtbbox \mid \gtbbox \in \epopa \;\land\; \phi_{e}(\gtbbox) > \lambda_{e} \cdot \lambda_{a} \}$ and
$\ambpop^C = \{ \gtbbox \mid \gtbbox \in \crpopa \;\land\; \phi_{c}(\gtbbox) > \lambda_{c} \cdot \lambda_{a} \}$.



Ground truth bounding boxes with ambiguous occlusion are defined as $\ambpop = \ambpop^E \cup \ambpop^C$. 
Based on that, the set of environmentally occluded ground truth bounding boxes is reduced to $\epopb = \epopa \setminus \ambpop^E$ and the set of crowd occluded ground truth bounding boxes is $\crpopb = \crpopa \setminus \ambpop^C$.
For our experiments, we empirically set $\lambda_a = 0.75$.

\paragraph{Foreground and Background}

\begin{figure*}[ht]
\centering
\includegraphics[width=\linewidth]{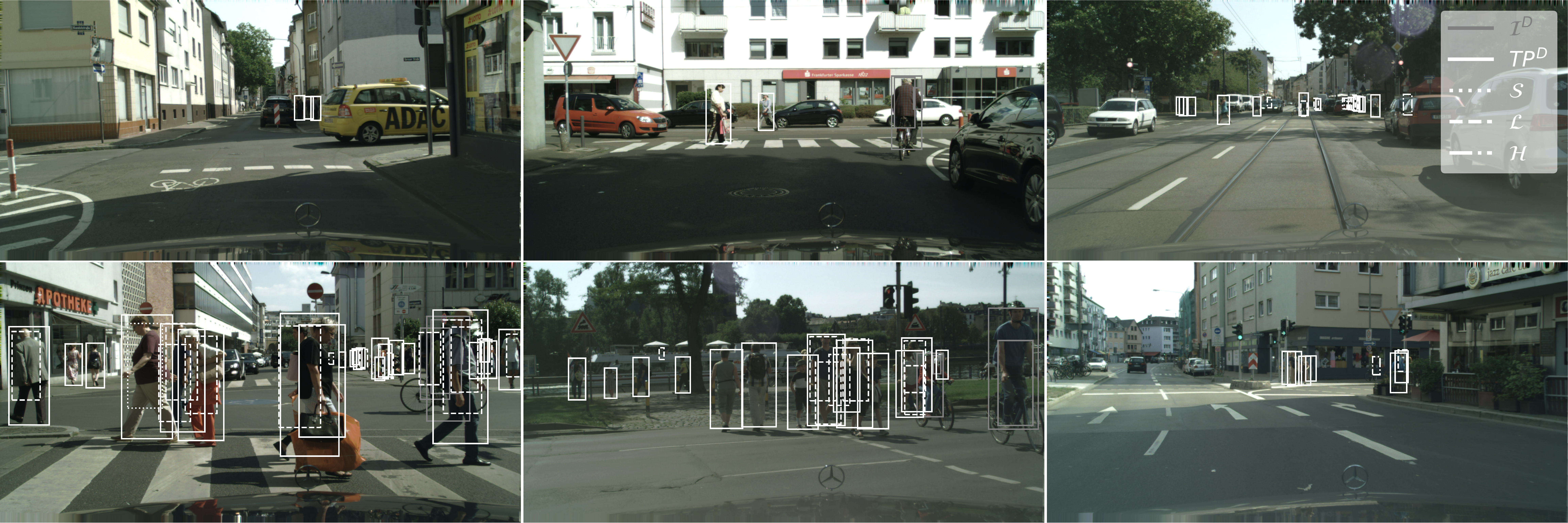}
\caption{Categories for detection bounding boxes: true positives $\text{TP}^D$ (solid), ghost detections $\gpop$ (dash dotted), localization errors $\lpop$ (dashed) scale errors $\spop$ (dotted) and \textcolor{gray}{ignored} detections $\ignd$ (solid).}
\label{fig:gt_err_cat_dt}
\end{figure*}

After defining occluded ground truth bounding boxes as $\occlpop = \epopb \cup \crpopb \cup \ambpop$, the clearly visible bounding boxes are given by $\vispop = \allgtbbox \setminus \occlpop$. 
By applying a height threshold $\lambda_{f}$, we can further divide $\vispop$ into foreground $\forepop$ or background $\backpop$. First, we define the foreground $\forepop = \{ \gtbbox \mid \gtbbox \in \vispop \;\land\; \text{height}(\gtbbox) \ge \lambda_{f} \}$.
Then, $\fore = \fn \cap \forepop$ defines errors in the foreground. 
In order to define a reasonable $\lambda_{f}$, the braking distance of an automated emergency braking $d_{\text{AEB}}$ is defined as 
$d_{\text{AEB}} = d_{s} + d_{v} + \left \lceil{\frac{v^2}{2 \cdot \mu \cdot g}}\right \rceil + \left \lceil{v \cdot t_{proc}}\right \rceil$.

\begin{table}[ht]
\centering
\begin{tabular}{l|l|l}
\toprule
Parameter & Value & Description \\
\midrule
\midrule
$h$         & $1.7\:m$                  & Pedestrian height \\
$t_{proc}$  & $0.4 \: sec$               & Processing time \\
$\mu$       & $0.3$                     & Friction coefficient \\
$v$         & $8.33 \:\frac{m}{s}$       & Velocity \\
$g$         & $9.81 \:\frac{m}{s^2}$    & Gravitational constant \\
$d_s$       & $2\:m$                    & Added distance \\
$d_v$       & $4\:m$                    & Distance from rear axis to front \\
\bottomrule
\end{tabular}
\caption{Parameters for the simplified braking distance calculation of an automated emergency braking $d_{\text{AEB}}$ with $30 \:\frac{km}{h}$.}
\label{tab:aeb}
\end{table}

Applying the parameters shown in Table \ref{tab:aeb}, the separating distance is $d_{\text{AEB}} = 22\:m$. 
Based on camera calibration parameters of Cityscapes, the corresponding pixel height $\lambda_{f}$ is approximately $190 \text{ pixels}$.
Finally, the background is specified as $\backpop = \vispop \setminus \forepop$ and potential background errors are defined as: $\back = \fn \cap \backpop$. 


Due to highly crowd-occluded pedestrians that introduce doubtful false negatives into the evaluation, we relax the matching strategy in Equation \ref{eq:tp}: 
For all $\gtbbox \in \forepop \cup \backpop$, we see $\gtbbox$ as a true positive if there exists a detection with $\iou > 0.5$ irrespective of another $\gtbbox \in \crpopb$ that could be matched with a higher $\iou$.

\subsection{Error Categories for False Positives}

A detection bounding box $\dtbbox \in \fp$ is a false positive. 
We argue that false positives that coincide with pedestrian crowds do not disrupt the operation of an ADS as much as unrelated and random false positives.
Hence, we propose three error categories with respect to false positives in order to identify the most disruptive. 
For examples see Figure \ref{fig:gt_err_cat_dt}.

\paragraph{Scale Errors}

This category includes detections that fail only with respect to the scale of the bounding box.
Let $\cx(\bbox), \cy(\bbox)$ give the x- and y-center coordinates of any bounding box $\bbox$ and $\lambda_{o}$ the maximum permitted center offset. 
The predicate that states whether the center of $\dtbbox$ is aligned with $\gtbbox$, is defined as:

\begin{equation}
    \label{eq:center-aligned}
    \begin{split}
        &\caligned(\gtbbox, \dtbbox) \iff |\cx(\gtbbox) - \cx(\dtbbox)| \le \lambda_{o} w(\gtbbox) \\
        & \hspace{6em}\land |\cy(\gtbbox) - \cy(\dtbbox)| \le \lambda_{o} h(\gtbbox)
    \end{split}
\end{equation}

For our experiments, we empirically set $\lambda_o =0.2$. Scale errors are defined as 
$\s = \{ \dtbbox \mid \dtbbox \in \fp \;\land\; \exists \gtbbox \in \allgtbbox : \caligned(\gtbbox, \dtbbox) \}$.

\paragraph{Localization Errors}

Holds all false positives that fall in close proximity to a $\gtbbox$, but the detection can not be matched and is not a scale error.
Localization errors are defined as 
$\lc = \{ \dtbbox \mid \dtbbox \in \left( \fp \setminus \s \right) \;\land\; \exists \gtbbox \in \allgtbbox : \iou(\gtbbox, \dtbbox) \ge \lambda_{i}\}$.
For our experiments, we empirically set $\lambda_i = 0.25$.

\paragraph{Ghost Detections}

Inspired by a term from automotive radar systems \cite{scheiner}, we define ghost detections as $\g = \fp \setminus \left( \s \;\cup\; \lc \right)$. 
Detections in this category are random and unrelated to the presence of pedestrians.
Thus, these are strongly disruptive that severely impact the operation of an ADS.

\subsection{Filtered Log-Average Miss Rate}

\begin{table*}[ht]
\centering
\begin{tabular}{l|ccc|cc|cc|cc|cc}
\toprule
\multirow{3}{*}{Feature Extraction} & 
\multicolumn{3}{c|}{\textbf{$\text{LAMR}$}} & 
\multicolumn{4}{c|}{\textbf{$\text{FLAMR}_{P}$}} & 
\multicolumn{4}{c}{\textbf{$\text{FLAMR}_{P}^{\gpop}$}} \\
& \multicolumn{3}{c|}{reasonable} &
\multicolumn{2}{c|}{$\forepop$} & 
\multicolumn{2}{c|}{$\backpop$} &
\multicolumn{2}{c|}{$\forepop$} & 
\multicolumn{2}{c}{$\backpop$} \\
& best & $\mu$ & $CI_{0.95}$ &
$\mu$ & $CI_{0.95}$ & 
$\mu$ & $CI_{0.95}$ & 
$\mu$ & $CI_{0.95}$ & 
$\mu$ & $CI_{0.95}$ \\
\midrule
\midrule
FPN-ResNet-50 & 10.9 & 11.6 & [11.2, 12.1] & 4.5 & [4.2, 4.9] & 12.4 & [11.7, 13.1] & 1.9 & [1.2, 2.5] & 6.8 & [6.3, 7.3]  \\
CSP-ResNet-50 & 10.6 & 11.0 & [10.7, 11.3] & 5.2 & [4.8, 5.5] & 11.2 & [10.8, 11.6] & 2.2 & [1.9, 2.4] & 6.3 & [6.2, 6.5]  \\
MDLA-UP-34 & 9.6 & 10.5 & [10.1, 10.8] & 4.7 & [4.2, 5.2] & 10.4 & [10.0, 10.8] & 2.8 & [2.5, 3.1] & 6.6 & [6.3, 6.9]  \\
BGC-HRNet-w32 & 8.8 & 9.1 & [9.0, 9.2] & 3.8 & [3.2, 4.4] & 9.0 & [8.7, 9.4] & 1.6 & [1.2, 2.0] & 5.6 & [5.3, 5.8]  \\
\bottomrule
\end{tabular}
\caption{Results of our experiments over different metrics.}
\label{tab:flamr}
\end{table*}

In the following, we derive metrics to measure the performance of a PD over the proposed error categories. Table \ref{tab:cat} shows the number of ground truth bounding boxes for each category. 
The filtered miss rate $\mrp$ accounts for ground truth bounding boxes with $\errorpop \in \{ \forepop, \backpop, \epopb, \crpopb, \ambpop \}$:

\begin{equation}
    \mrp = \frac{|\fn \cap \errorpop|}{|\tp \cap \errorpop| + |\fn \cap \errorpop|}
\end{equation}

\begin{table}[ht]
\centering
\begin{tabular}{l|ccccc}
\toprule
     \textbf{Subset} & $\forepop$ & $\backpop$ & $\epopb$ & $\crpopb$ & $\ambpop$ \\
     \midrule
     \textbf{Cardinality} & 348 & 1269 & 364 & 438 & 130 \\
\bottomrule
\end{tabular}
\caption{Allocation of ground truth bounding boxes for the CityPersons validation dataset.}
\label{tab:cat}
\end{table}

\paragraph{False Positives per Image}

With reference to the $\lamr$, the filtered log-average miss rate ($\flamr$) is defined as

\begin{equation}
\label{eq:flamr}
    \flamr = \exp \left ( \frac{1}{|\cset|} \sum_{c \in \cset} \log 
    \mrS_{\errorpop}(c) \right )
\end{equation}

Here, $\cset$ is a set of confidence levels that correspond to the nine pre-defined $\fppi$ values for calculating the $\lamr$:

\begin{equation}
    \cset = \left\{ \underset{\fppi \le f}{\argmax} \; \fppi \mid f \in F \right\}
\label{eq:cset}
\end{equation}

with $F = \{10^{-2}, 10^{-1.78}, \ldots, 10^0\}$ and $|F|=9$.

\paragraph{Ghost Detections per Image}

Since not all false positives are equally disruptive, we propose to focus on the number of ghost detections per image $\gdpi(c) = \frac{1}{N} |\g|$. Based on $\gdpi(c)$ and Equation \ref{eq:cset}, we denote the set of confidence levels for ghost detections as $\chset$. 
The filtered log-average miss rate with respect to ghost detections ($\flamrh$) is defined as:\looseness=-1

\begin{equation}
    \flamrh = \exp \left ( \frac{1}{|\chset|} \sum_{c \in \chset} \log 
    \mrS_{\errorpop}(c) \right )
\end{equation}


\subsection{Upper Bound for Confidence Threshold}

From a safety perspective, we are interested in finding an operating point for a PD where no safety-critical pedestrian is missed.
It is still open to what extent this requirement can be relaxed for DNNs for object tracking.
In this work, we are focused on single images and define a safety-critical pedestrian as any pedestrian who is in the foreground. 
Furthermore, we assign the operating point to a confidence threshold that must be determined post-hoc to PD training.
We define the confidence threshold $c \in [ 0, c^*_{\forepop} ]$ and the upper bound $c^*_{\forepop}$ as the confidence threshold with the lowest miss rate for foreground $\forepop$: $c^*_{\forepop} = \argmin \text{MR}_{\forepop}(c)$.
If $\text{MR}_{\forepop}(c^*_{\forepop}) = 0$ holds for a given validation dataset, it can be ensured that every safety-critical pedestrian in the foreground is correctly detected.
Lowering the confidence threshold so that $c < c^*_{\forepop}$ may improve performance for other error categories, but is not capable of causing foreground errors.
Consequently, we see $c^*_{\forepop}$ as a reasonable choice for an operating point.
The corresponding amount of ghost detections per image is given by $\text{GDPI}(c^*_{\forepop})$.

\section{Results}
\label{sec:results}

In total, we trained and analyze results for 44 PDs i.e. 11 PDs for each of the four different feature extractions and backbones.
PDs with the same feature extraction also differ since the randomly initialized AL parameters in the feature extraction and perception heads change for every training run. 
We report confidence intervals ($CI_{0.95}$), using a student's t-distribution due to the small sample size.
This accounts for randomness and improves transparency, although satisfactory sample sizes are difficult when working with large DNNs.

\subsection{Log-Average Miss Rate}

LAMR scores for the reasonable subset of the CityPersons validation dataset are reported in Table \ref{tab:flamr}. 
BGC-HRNet-w32 has the best LAMR performance, which confirms the reported benchmarks listed in Table \ref{tab:lamr_literature}. 
In summary, our experiments show overlapping $CI_{0.95}$, indicating that randomness in initialization and training influences performance.
Interestingly, our PD with BGC-HRNet-w32 as backbone achieves a score very similar to the results reported for BGCNet \cite{li_box_2020} despite using a simpler architecture that does not employ box-guided convolutions.

\subsection{Bias of Reasonable Subset}

Compared to the $\lamrR$ scores on the reasonable subset, we see a corresponding order of the $\flamr$ scores for background in Table \ref{tab:flamr}.
This indicates that the reasonable subset holds a vast amount of smaller pedestrians in the background.
BGC-HRNet-w32 performs best for all performance metrics and subsets.
In contrast, $\flamrf$ scores contradict the $\lamrR$ results with a different ranking of PDs and strongly overlapping $CI_{0.95}$.
The inherent bias of the $\lamrR$ evaluation leads to underestimation of the true potential of certain feature extractions and backbones for the highly safety-critical foreground category.

\begin{figure}[ht]
\centering
\includegraphics[width=\linewidth]{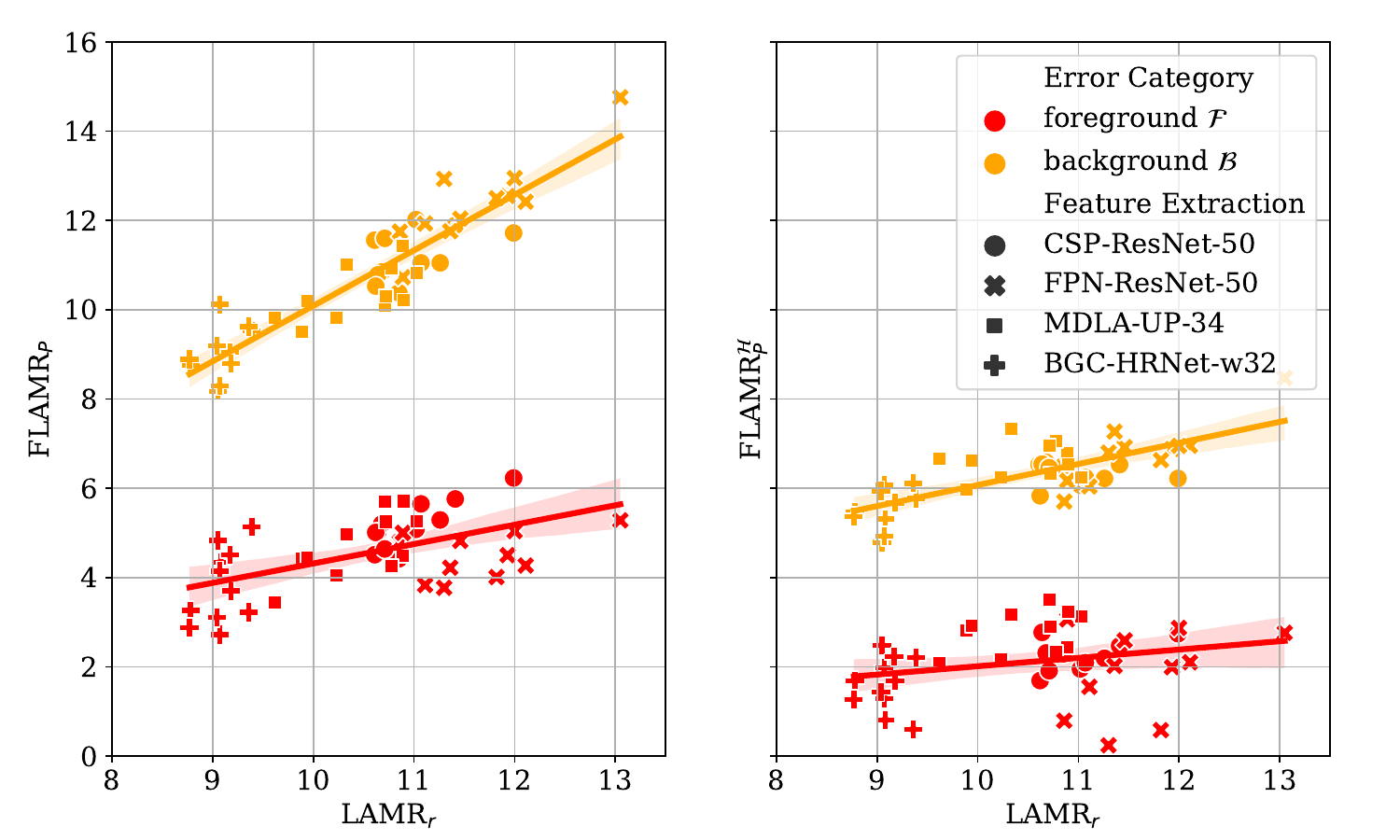}
\caption{LAMR scores for the reasonable subset compared to the filtered log-average miss rate with and w/o respect to ghost detections ($\flamr$, $\flamrh$).}
\label{fig:lamr_flamr}
\end{figure}

Figure \ref{fig:lamr_flamr} analyzes the dependence of $\lamrR$ scores and $\flamr$ and $\flamrh$ for foreground and background.
Whereas $\flamr$ scores are strongly correlated with $\lamrR$ scores in the background, the dependence in the foreground is lower. 
Our analysis shows that the reasonable subset is dominated by pedestrians in the background, which are less safety-critical.

\subsection{Application-Oriented Evaluation}

Evaluating the miss rate in foreground $\forepop$ and background $\backpop$ while only considering ghost detections per image ($\gdpi$) combines the evaluation of opposed critical cases: Missing a safety-critical pedestrian or predicting non-existing pedestrians.
The evaluation of the best run in terms of $\lamrR$ and $\flamr$ in Table \ref{tab:flamr} shows BGC-HRNet-w32 as the far superior choice.
However, $\flamrhf$-scores based on our systematic error categories reveal that performance of FPN-ResNet-50 is comparable by achieving the same lower bound of $CI_{0.95}$.
This is contradictory to the fact that BGC-HRNet-w32 outperforms FPN-ResNet-50 by nearly $2\%$ in $\lamrR$.
We observe the reversed effect for MDLA-UP-34 which performs second-best in $\lamrR$ but achieves the worst result for $\flamrhf$.

Figure \ref{fig:lamr_flamr} shows how the proposed focus on ghost detections almost resolves the weak dependence between $\flamrh$ to $\lamrR$ in the foreground.
Hence, the $\flamrh$ effectively measures performance differently and considers factors that are ignored by the $\lamrR$.
The results show that PDs optimized for $\lamrR$ do not necessarily perform best with respect to $\flamr$ or $\flamrh$.
Controversially, there are PDs (with FPN-ResNet-50) that have a lower $\flamrhf$ score despite a much higher $\lamrR$ score.
These models have a lower miss rate for pedestrians in the foreground and produce fewer ghost detections per image.
Thus, the $\lamrR$ for the reasonable subset has limits in terms of an application-oriented evaluation.
The problem arises from the training strategy for PDs that is not focused on safety. 
The selection of the best performing checkpoint in terms of the $\lamrR$ is disconnected from the evaluation of safety-critical pedestrians.

Based on the large deviations between $\flamrh$ to $\lamrR$, we conclude that $\flamrh$ introduces a new application-oriented perspective for the evaluation of DNNs for pedestrian detection.
This conclusion seems reasonable due to the systematic categorization of errors. 
Here, safety-critical pedestrians are identified as the complement of highly occluded pedestrians and distant pedestrians.

\subsection{Operating Point}

\begin{figure}[ht]
\centering
\includegraphics[width=\linewidth]{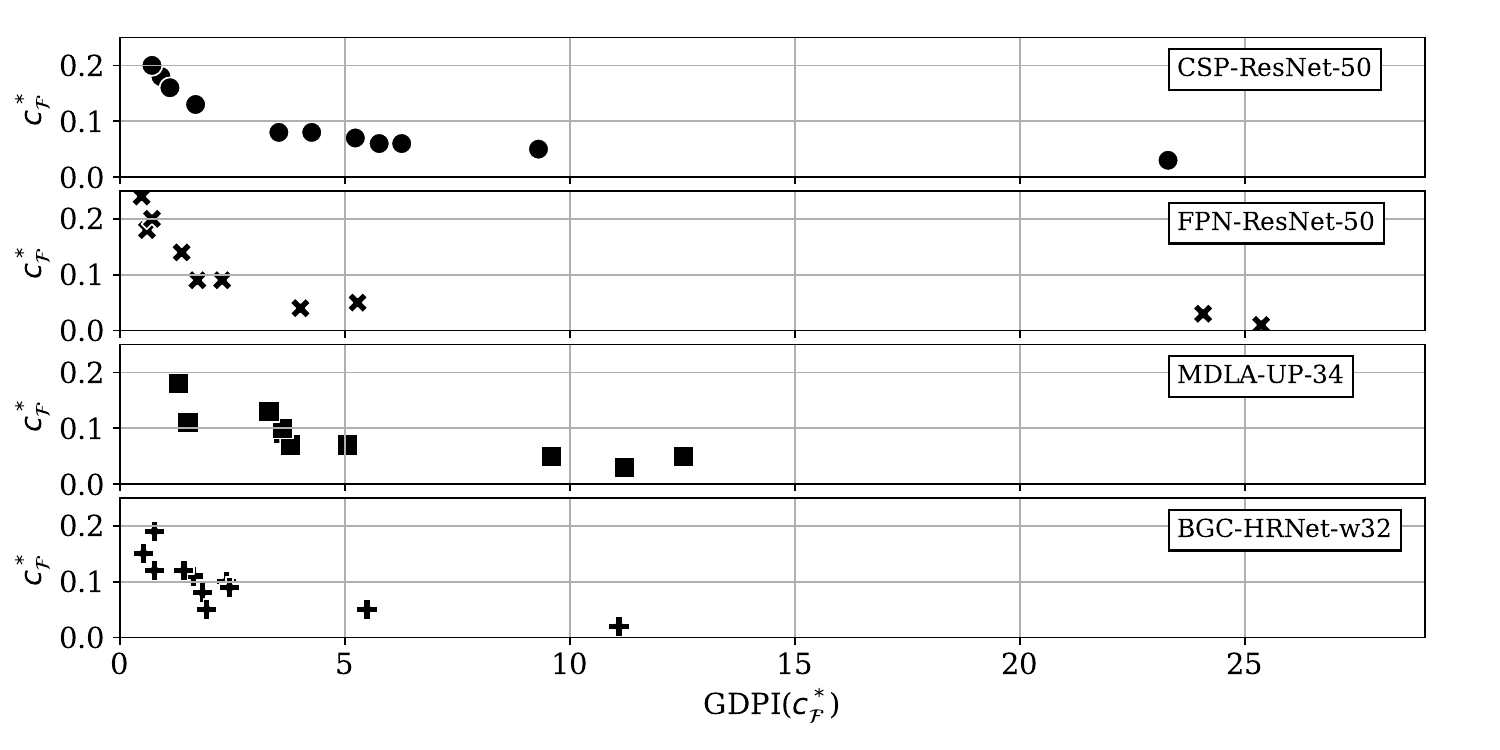}
\caption{Upper bound for confidence threshold ($c^*_{\forepop}$) for all tested PDs with the required amount of $\gdpi$.}
\label{fig:gdpi_vs_c_F}
\end{figure}

Towards an application-oriented analysis of DNNs for pedestrian detection, we determine upper bounds on the confidence threshold $c^*_{\forepop}$ as operating points for individual PDs. 
Results can be seen in Figure \ref{fig:gdpi_vs_c_F}.
We see that between training runs of PDs with the same feature extraction, the upper bound of the confidence threshold $c^*_{\forepop}$ and the required $\text{GDPI}(c^*_{\forepop})$ vary greatly.
Thus, operating points must be determined individually for the PDs and cannot be specified in general for a particular feature extraction.  

Furthermore, our evaluation shows that not every PD is capable of detecting every pedestrian in the foreground with a confidence threshold of 0.01. 
This means that there are PDs with $\text{MR}_{\mathcal{F}}(0.01) \neq 0$ (CSP-ResNet-50: 2, FPN-ResNet-50: 4, MDLA-UP-34: 0, BGC-HRNet-w32: 1).
In general, foreground pedestrians are missed with a maximum of $\text{MR}_{\forepop}(c^*_{\forepop}) = 0.29\%$.
Up to this point, the subset-based evaluation of DNNs for pedestrian detection has largely focused on benchmarking.
Due to the limited informative value, it was difficult to derive guidelines for the application-oriented development process of DNNs.
We take the stance that aggregated performance metrics such as $\flamrh$ must be collated with metrics such as $\text{MR}_{\mathcal{F}}(c^*_{\forepop})$ and $\text{GDPI}(c^*_{\forepop})$.

\begin{figure}[ht]
\centering
\includegraphics[width=0.9\linewidth]{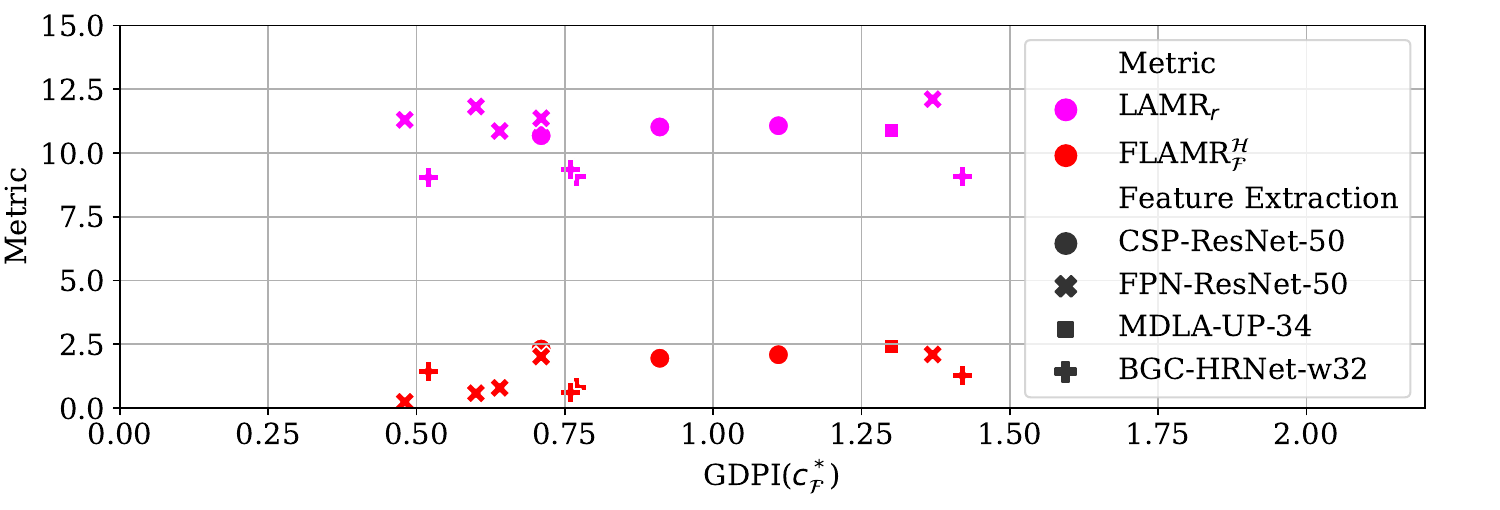}
\caption{$\lamrR$ and $\flamrh$ are independent of GDPI.}
\label{fig:gdpi}
\end{figure}

\begin{figure}[ht]
\centering
\includegraphics[width=\linewidth]{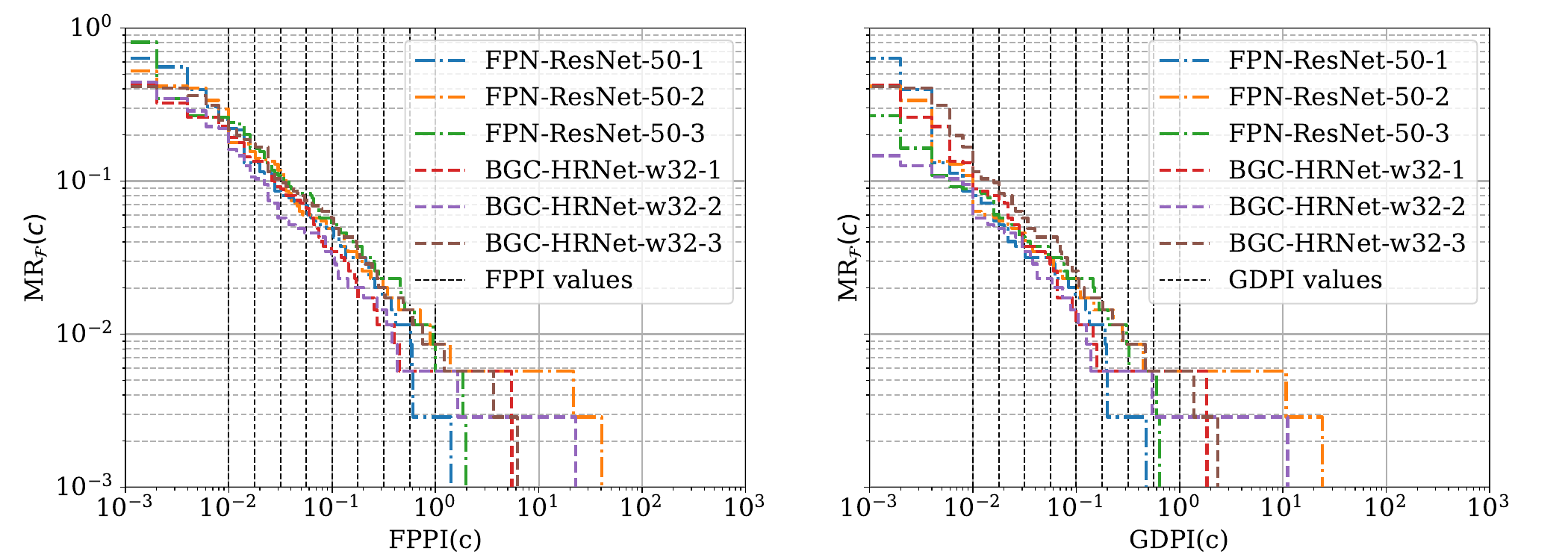}
\caption{Comparison of selected PDs for false positives per image ($\fppiS$, left) and ghost detections per image ($\gdpi$, right). The filtered miss rate $\text{MR}_{\mathcal{F}}(c)$ is calculated for pedestrians in the foreground $\forepop$.}
\label{fig:mr_fppi_curve}
\end{figure}

Figure \ref{fig:gdpi} shows that performance metrics ($\lamrR$ and $\flamrh$) are unrelated to the required number of ghost detections $\text{GDPI}(c^*_{\forepop})$. 
Surprisingly, FPN-ResNet-50 achieves with 0.48 the lowest value of $\text{GDPI}(c^*_{\forepop})$. 
The reason for the unexpected behavior can be seen in Figure \ref{fig:mr_fppi_curve}.
Although the sorted miss rate curves of the selected PDs are close in the middle range, they diverge the most in the head and tail ranges.
The vertical lines mark the common values for which the miss rate is averaged.
As a consequence, aggregated performance metrics such as $\lamrR$, $\flamr$ and $\flamrh$ average over multiple confidence thresholds and put less weight on safety-relevant ranges towards $c^*_{\forepop}$. 


\begin{figure}[ht]
\centering
\includegraphics[width=\linewidth]{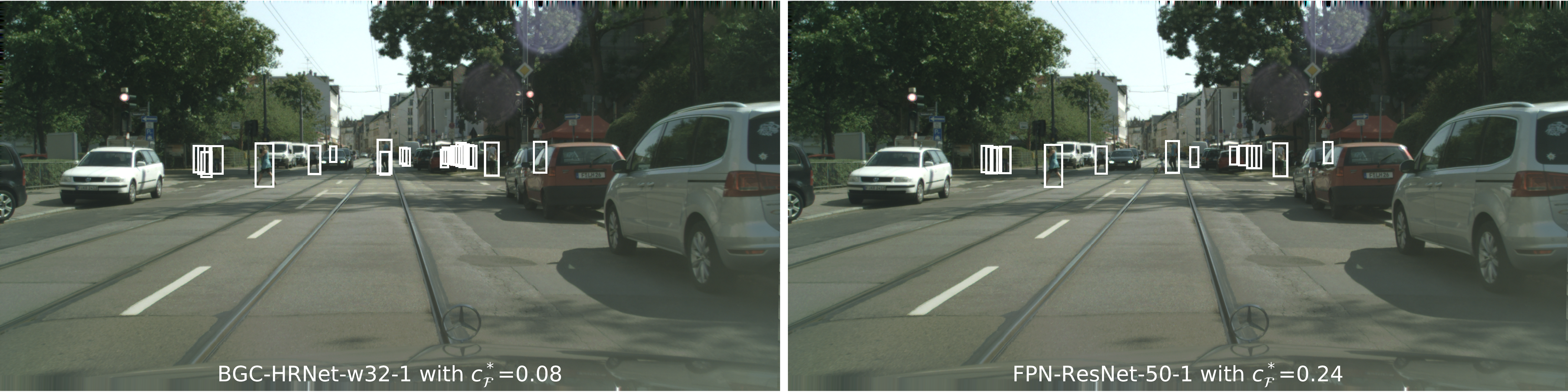}
\caption{Inference results for BGC-HRNet-w32 (first run, best $\lamrR$ score with 8.8\%) and FPN-ResNet50 (first run).}
\label{fig:cherry_pick}
\end{figure}

Since $c^*_{\forepop}$ determines an operating point for a PD, it can serve as a meaningful confidence threshold to visually assess inference samples (see Figure \ref{fig:cherry_pick}). 
This provides practitioners with a reliable basis of information and allows them to evaluate applicability more intuitively.

\section{Conclusion}


In this work, we propose a rule-based error categorization to evaluate the performance of a DNN for pedestrian detection.
Multiple disjoint categories for false negatives are defined in order to identify safety-critical errors in the foreground.
The distinction is based on three occlusion-related categories and the braking distance of an automated driving system.
We expect that the inclusion of depth information would improve the separation between foreground, background, occluding pedestrians and environment.
In future work, we would like to reevaluate the performance of DNNs specifically designed for the occlusion problem using our proposed error categories.
We identify three categories of false positives, with ghost detections being the most disruptive. 
For our experiments, we use a simple and generic framework to build DNNs for pedestrian detection.
In consequence, we train 44 DNNs based on four commonly used backbones, achieving state-of-the-art performance in terms of $\lamrR$.
The goal of our application-oriented evaluation is two-folded.
To account for safety-critical false negatives as well as disruptive false positives, we propose $\flamrhf$ as a new performance metric.
Finally, we determine an operating point as the confidence threshold where no pedestrian in the foreground is missed.
By revisiting and refining the current evaluation, we contribute to a safety-focused development process of DNNs for pedestrian detection. 

\section*{Acknowledgements}
The research leading to these results is funded by the German Federal Ministry for Economic Affairs and Energy within the project “Methoden und Maßnahmen zur Absicherung von KI basierten Wahrnehmungsfunktionen für das automatisierte Fahren (KI-Absicherung)".
The authors would like to thank the consortium for the successful cooperation.
The authors acknowledge support by the state of Baden-Württemberg through bwHPC.

\FloatBarrier
\bibliographystyle{ijcai22}
\bibliography{ijcai22}
\end{document}